\documentclass[11pt]{article}
\usepackage{emnlp2016}
\usepackage{times}
\usepackage{latexsym}
\usepackage{graphicx}
\usepackage{verbatim}

\emnlpfinalcopy % Uncomment this line for the final submission
\def\emnlppaperid{204} %  Enter the acl Paper ID here

\title{Building an Evaluation Scale using Item Response Theory}

\author{John P. Lalor$^1$, Hao Wu$^2$, Hong Yu$^{1,3}$ \\
	$^1$ University of Massachusetts, MA, USA \\
    $^2$ Boston College, MA, USA \\
    $^3$ Bedford VAMC and CHOIR, MA, USA \\
    {\tt lalor@cs.umass.edu, hao.wu.5@bc.edu, hong.yu@umassmed.edu}\\
}

\date{}

\begin{document}

\maketitle

\begin{abstract}
Evaluation of NLP methods requires testing against a previously vetted gold-standard test set and reporting standard metrics (accuracy/precision/recall/F1). 
The current assumption is that all items in a given test set are equal with regards to difficulty and discriminating power. 
We propose Item Response Theory (IRT) from psychometrics as an alternative means for gold-standard test-set generation and NLP system evaluation.
IRT is able to describe characteristics of individual items - their difficulty and discriminating power - and can account for these characteristics in its estimation of human intelligence or ability for an NLP task.
In this paper, we demonstrate IRT by generating a gold-standard test set for Recognizing Textual Entailment. 
By collecting a large number of human responses and fitting our IRT model, we show that our IRT model compares NLP systems with the performance in a human population and is able to provide more insight into system performance than standard evaluation metrics.
We show that a high accuracy score does not always imply a high IRT score, which depends on the item characteristics and the response pattern.\footnote{Data and code will be made available for download at https://people.cs.umass.edu/lalor/irt.html}

\end{abstract}

\section{Introduction}

Advances in artificial intelligence have made it possible to compare computer performance directly with human intelligence \cite{campbell2002deep,ferrucci2010building,silver2016mastering}. In most cases, a common approach to evaluating the performance of a new system is to compare it against an unseen gold-standard test dataset (GS items). Accuracy, recall, precision and F1 scores are commonly used to evaluate NLP applications. These metrics assume that GS items have equal weight for evaluating performance. However, individual items are different: some may be so hard that most/all NLP systems answer incorrectly; others may be so easy that every NLP system answers correctly. Neither item type provides meaningful information about the performance of an NLP system. Items that are answered incorrectly by some systems and correctly by others are useful for differentiating systems according to their individual characteristics. 

In this paper we introduce Item Response Theory (IRT) from psychometrics and demonstrate its application to evaluating NLP systems. IRT is a theory of evaluation for characterizing test items and estimating human ability from their performance on such tests. IRT assumes that individual test questions (referred to as ``items'' in IRT) have unique characteristics such as difficulty and discriminating power. These characteristics can be identified by fitting a joint model of human ability and item characteristics to human response patterns to the test items. Items that do not fit the model are removed and the remaining items can be considered a scale to evaluate performance. IRT assumes that the probability of a correct answer is associated with both item characteristics and individual ability, and therefore a collection of items of varying characteristics can determine an individual's overall ability.

Our aim is to build an intelligent evaluation metric to measure performance for NLP tasks. With IRT we can identify an appropriate set of items to measure ability in relation to the overall human population as scored by an IRT model. This process serves two purposes: (i) to identify individual items appropriate for a test set that measures ability on a particular task, and (ii) to use the resulting set of items as an evaluation set in its own right, to measure the ability of future subjects (or NLP models) for the same task. These evaluation sets can measure the ability of an NLP system with a small number of items, leaving a larger percentage of a dataset for training. 

Our contributions are as follows: First, we introduce IRT and describe its benefits and methodology. Second, we apply IRT to Recognizing Textual Entailment (RTE) and show that evaluation sets consisting of a small number of sampled items can provide meaningful information about the RTE task. Our IRT analyses show that different items exhibit varying degrees of difficulty and discrimination power and that high accuracy does not always translate to high scores in relation to human performance. By incorporating IRT, we can learn more about dataset items and move past treating each test case as equal. Using IRT as an evaluation metric allows us to compare NLP systems directly to the performance of humans.   

\section{Background and Related Work}
\subsection{Item Response Theory}
\label{ssec:irtbackground}

IRT is one of the most widely used methodologies in psychometrics for scale construction and evaluation. It is typically used to analyze human responses (graded as right or wrong) to a set of questions (called ``items''). With IRT individual ability and item characteristics are jointly modeled to predict performance \cite{baker_item_2004}. This statistical model makes the following assumptions: (a) Individuals differ from each other on an unobserved latent trait dimension (called ``ability'' or ``factor''); (b) The probability of correctly answering an item is a function of the person's ability. This function is called the item characteristic curve (ICC) and involves item characteristics as parameters; (c) Responses to different items are independent of each other for a given ability level of the person (``local independence assumption''); (d) Responses from different individuals are independent of each other.

More formally, if we let $j$ be an individual, $i$ be an item, and $\theta_j$ be the latent ability trait of individual $j$, then the probability that individual $j$ answers item $i$ correctly can be modeled as:
\begin{equation} \label{eq:3pl}
  p_{ij} (\theta_j )= c_i +  \frac{1- c_i}{1+ e^{-a_i (\theta_j-b_i)} }
\end{equation}
where $a_i$, $b_i$, and $c_i$ are item parameters: $a_i$ (the slope or discrimination parameter) is related to the steepness of the curve, $b_i$ (the difficulty parameter) is the level of ability that produces a chance of correct response equal to the average of the upper and lower asymptotes, and $c_i$ (the guessing parameter) is the lower asymptote of the ICC and the probability of guessing correctly. Equation \ref{eq:3pl} is referred to as the three-parameter logistic (3PL) IRT model. A two-parameter logistic (2PL) IRT model assumes that the guessing parameter $c_i$ is 0.

\begin{figure}
\centering
\includegraphics[width=\columnwidth]{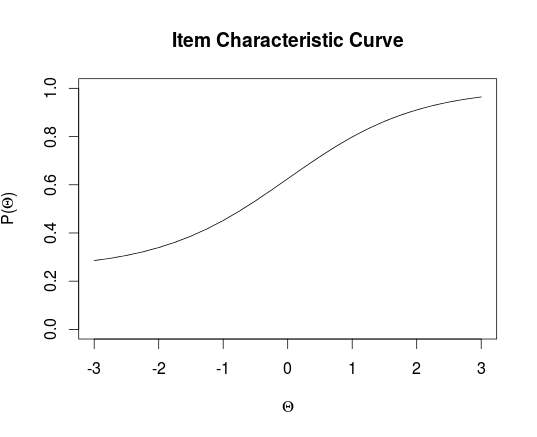}
\caption{Example ICC for a 3PL model with the following parameters: $a=1.0$, $b=0.0$, $c=0.25$.}
\label{fig:irtexample}
\end{figure}

Figure \ref{fig:irtexample} shows an ICC of a 3PL model. 
The ICC for a good item will look like a sigmoid plot, and should exhibit a relatively steep increasing ICC between ability levels $-3$ and 3, where most people are located, in order to have appropriate power to differentiate different levels of ability. We have described a one factor IRT model where ability is uni-dimensional. Multi-factor IRT models would involve two or more latent trait dimensions and will not be elaborated here.

To identify the number of factors in an IRT model, the polychoric correlation matrix of the items is calculated and its ordered eigenvalues are plotted. The number of factors is suggested by the number of large eigenvalues. It can be further established by fitting (see below) and comparing IRT models with different numbers of factors. Such comparison may use model selection indices such as Akaike Information Criterion (AIC) and Conditional Bayesian Information Criterion (CBIC) and should also take into account the interpretablility of the loading pattern that links items to factors.

An IRT model can be fit to data with the marginal maximum likelihood method through an EM algorithm \cite{bock1981marginal}. The marginal likelihood function is the probability to observe the current response patterns as a function of the item parameters with the persons' ability parameters integrated out as random effects. This function is maximized to produce estimates of the item parameters. For IRT models with more than one factor, the slope parameters (i.e. loadings) that relate items and factors must be properly rotated \cite{browne2001overview} before they can be interpreted. Given the estimated item parameters, Bayesian estimates of the individual person's ability parameters are obtained with the standard normal prior distribution.

After determining the number of factors and fitting the model, the local independence assumption can be checked using the residuals of marginal responses of item pairs \cite{chen_local_1997} and the fit of the ICC for each item can be checked with item fit statistics \cite{orlando_likelihood-based_2000} to determine whether an item should be retained or removed. If both tests are passed and all items have proper discrimination power, then the set of items is considered a calibrated measurement scale and the estimated item parameters can be further used to estimate an individual person's ability level.

IRT accounts for differences among items when estimating a person's ability. In addition, ability estimates from IRT are on the ability scale of the population used to estimate item parameters. For example, an estimated ability of 1.2 can be interpreted as 1.2 standard deviations above the average ability in this population. The traditional total number of correct responses generally does not have such quantitative meaning.

IRT has been widely used in educational testing. For example, it plays an instrumental role in the construction, evaluation, or scoring of standardized tests such as the Test of English as a Foreign Language (TOEFL), Graduate Record Examinations (GRE) and the SAT college admissions standardized test.

\subsubsection{IRT Terminology}

Here we outline common IRT terminology in terms of RTE. An \textit{item} refers to a pair of sentences to which humans or NLP systems assign a label (entailment, contradiction, or neutral). A set of responses to all items (each graded as correct or incorrect) is a \textit{response pattern}. An \textit{evaluation scale} is a test set of items to be administered to an NLP system and assigns an \textit{ability score} (or \textit{theta score}) to the system as its performance. 

\subsection{Recognizing Textual Entailment}
\label{ssec:rte}

RTE was introduced to standardize the challenge of accounting for semantic variation when building models for a number of NLP applications \cite{dagan_pascal_2006}. RTE defines a directional relationship between a pair of sentences, the text (T) and the hypothesis (H). T entails H if a human that has read T would infer that H is true. If a human would infer that H is false, then H contradicts T. If the two sentences are unrelated, then the pair are said to be neutral. Table \ref{tab:examples} shows examples of T-H pairs and their respective classifications. Recent state-of-the-art systems for RTE require a large amount of feature engineering and specialization to achieve high performance \cite{beltagy_representing_2015,lai_illinois-lh:_2014,jimenez_unal-nlp:_2014}.

\begin{table*}[th]
\small
\centering
\begin{tabular*}{\textwidth}{|p{8.5cm} @{\extracolsep{\fill}}p{4.5cm}l|}
\hline \bf Text & \bf Hypothesis & \bf Label \\ \hline
\bf Retained - 4GS & & \\ \hline
1. A toddler playing with a toy car next to a dog & A toddler plays with toy cars while his dog sleeps & Neutral \\
2. People were watching the tournament in the stadium & The people are sitting outside on the grass & Contradiction \\
\hline \bf Retained - 5GS & & \\ \hline
3. A person is shoveling snow & It rained today & Contradiction \\
4 Two girls on a bridge dancing with the city skyline in the background & The girls are sisters. & Neutral \\
5. A woman is kneeling on the ground taking a photograph & A picture is being snapped & Entailment \\
\hline \bf Removed - 4GS & & \\ \hline
6. Two men and one woman are dressed in costume hats & The people are swingers & Neutral \\
7. Man sweeping trash outside a large statue & A man is on vacation & Contradiction \\
8. A couple is back to back in formal attire & Two people are facing away from each other & Entailment \\
9. A man on stilts in a purple, yellow and white costume & A man is performing on stilts & Entailment \\
\hline \bf Removed - 5GS & & \\ \hline
10. A group of soccer players are grabbing onto each other as they go for the ball & A group of football players are playing a game & Contradiction \\
11. Football players stand at the line of scrimmage & The players are in uniform & Neutral \\
12. Man in uniform waiting on a wall & Near a wall, a man in uniform is waiting & Entailment \\
\hline
\end{tabular*}
\caption{Examples of retained \& removed sentence pairs. The selection is not based on right/wrong labels but based on IRT model fitting and item elimination process. Note that no 4GS entailment items were retained (Section \ref{ssec:irtevaluation})}\label{tab:examples}
\end{table*}

A number of gold-standard datasets are available for RTE \cite{marelli_sick_2014,young_image_2014,levy_focused_2014}. We consider the Stanford Natural Language Inference (SNLI) dataset \cite{bowman_large_2015}. SNLI examples were obtained using only human-generated sentences with Amazon Mechanical Turk (AMT) to mitigate the problem of poor data that was being used to build models for RTE. In addition, SNLI included a quality control assessment of a sampled portion of the dataset (about 10\%, 56,951 sentence pairs). This data was provided to 4 additional AMT users to provide labels (entailment, contradiction, neutral) for the sentence pairs. If at least 3 of the 5 annotators (the original annotator and 4 additional annotators) agreed on a label the item was retained. Most of the items (98\%) received a gold-standard label. Specifics of SNLI generation are at \newcite{bowman_large_2015}. 

\subsection{Related Work}

To identify low-quality annotators (\textit{spammers}), \newcite{hovy2013learning} modeled annotator responses, either answering correctly or guessing, as a random variable with a guessing parameter varying only across annotators. \newcite{passonneau2014benefits} used the model of \newcite{dawid1979maximum} in which an annotator's response depends on both the true label and the annotator. In both models an annotator's response depends on an item only through its correct label. In contrast, IRT assumes a more sophisticated response mechanism involving both annotator qualities and item characteristics. To our knowledge we are the first to introduce IRT to NLP and to create a gold standard with the intention of comparing NLP applications to human intelligence.

\begin{comment}
\newcite{munro2010crowdsourcing} evaluate the quality of data generated for linguistics with AMT, and recreate classic linguistic studies and provide evaluation metrics for the obtained data. They compare crowd-generated data with controlled experiments, where in this study we use the crowd to identify dataset items for a discriminating test set for future evaluations.
\end{comment}

\newcite{bruce1999recognizing} analyze patterns of agreement between annotators in a case-study sentence categorization task, and use a latent-trait model to identify true labels. That work uses 4 annotators at varying levels of expertise and does not consider the discriminating power of dataset items.

Current gold-standard dataset generation methods include web crawling \cite{guo_linking_2013}, automatic and semi-automatic generation \cite{an_automatic_2003}, and expert \cite{roller_held-out_2015} and non-expert human annotation \cite{bowman_large_2015,wiebe_development_1999}. In each case validation is required to ensure that the data collected is appropriate and usable for the required task. Automatically generated data can be refined with visual inspection or post-collection processing. Human annotated data usually involves more than one annotator, so that comparison metrics such as Cohen's or Fleiss' $\kappa$ can be used to determine how much they agree. Disagreements between annotators are resolved by researcher intervention or by majority vote.

\section{Methods}

We collected and evaluated a random selection from the SNLI RTE dataset ($GS_{RTE}$) to build our IRT models. We first randomly selected a subset of $GS_{RTE}$, and then used the sample in an AMT Human Intelligence Task (HIT) to collect more labels for each text-hypothesis pair. We then applied IRT to evaluate the quality of the examples and used the final IRT models to create evaluation sets ($GS_{IRT}$) to measure ability for RTE.

\subsection{Item Selection}
\label{ssec:itemselection}

For our evaluation we looked at two sets of data: sentence-pairs selected from SNLI where 4 out of 5 annotators agreed on the gold-standard label (referred to as 4GS), and sentence-pairs where 5 out of 5 annotators agreed on the gold-standard label (referred to as 5GS). We make the assumption for our analysis that the 4GS items are harder than the 5GS items due to the fact that there was not a unanimous decision regarding the gold-standard label.

We selected the subset of $GS_{RTE}$ to use as an examination set in 4GS and 5GS according to the following steps: (1) Identify all ``quality-control'' items from $GS_{RTE}$ (i.e. items where 5 annotators provided labels, see \S \ref{ssec:rte}), (2) Identify items in this section of the data where 4 of the 5 annotators agreed on the eventual gold label (to be selected from for 4GS) and 5 of the 5 annotators agreed on the gold standard label (to be selected from for 5GS), (3) Randomly select 30 entailment sentence pairs, 30 neutral pairs, and 30 contradiction pairs from those items where 4 of 5 annotators agreed on the gold label (4GS) and those items where 5 of 5 annotators agreed on the gold label (5GS) to obtain two sets of 90 sentence pairs.

90 sentence pairs for 4GS and 5GS were sampled so that the annotation task (supplying 90 labels) could be completed in a reasonably short amount of time during which users remained engaged. We selected items from 4GS and 5GS because both groups are considered high quality for RTE. We evaluated the selected 180 sentence pairs using the model provided with the original dataset \cite{bowman_large_2015} and found that accuracy scores were similar compared to performance on the SNLI test set.

\subsection{AMT Annotation}

For consistency we designed our AMT HIT to match the process used to validate the SNLI quality control items \cite{bowman_large_2015} and to generate labels for the SICK RTE dataset \cite{marelli_sick_2014}. Each AMT user was shown 90 premise-hypothesis pairs (either the full 5GS or 4GS set) one pair at a time, and was asked to choose the appropriate label for each. Each user was presented with the full set, as opposed to one-label subsets (e.g. just the entailment pairs) in order to avoid a user simply answering with the same label for each item.

For each 90 sentence-pair set (5GS and 4GS), we collected annotations from 1000 AMT users, resulting in 1000 label annotations for each of the 180 sentence pairs. While there is no set standard for sample sizes in IRT models, this sample size satisfies the standards based on the non-central $\chi^2$ distribution \cite{maccallum1996power} used when comparing two multidimensional IRT models. This sample size is also appropriate for tests of item fit and local dependence that are based on small contingency tables. 

Only AMT users with approval ratings above 97\% were used to ensure that users were of a high quality. The task was only available to users located in the United States, as a proxy for identifying English speakers. Attention check questions were included in the HIT, to ensure that users were paying attention and answering to the best of their ability. Responses where the attention-check questions were answered incorrectly were removed. After removing individuals that failed the attention-check, we retained 976 labels for each example in the 4GS set and 983 labels for each example in the 5GS set. Average time spent for each task was roughly 30 minutes, a reasonable amount for AMT users.

\subsection{Statistical Analysis}
\label{ssec:stats}

Data collected for 4GS and 5GS were analyzed separately in order to evaluate the differences between ``easier'' items (5GS) and ``harder'' items (4GS), and to demonstrate the ability to show that theta score is consistent even if dataset difficulty varies. For both sets of items, the number of factors was identified by a plot of eigenvalues of the 90 x 90 tetrachoric correlation matrix and by a further comparison between IRT models with different number of factors. A target rotation \cite{browne2001overview} was used to identify a meaningful loading pattern that associates factors and items. Each factor could then be interpreted as the ability of a user to recognize the correct relationship between the sentence pairs associated with that factor (e.g. contradiction).

Once the different factors were associated with different sets of items, we built a unidimensional IRT model for each set of items associated with a single factor. We fit and compared one- and two-factor 3PL models to confirm our assumption and the unidimensional structure underlying these items, assuming the possible presence of guessing in people's responses. We further tested the guessing parameter of each item in the one factor 3PL model. If the guessing parameter was not significantly different from 0, a 2PL ICC was used for that particular item.

Once an appropriate model structure was determined, individual items were evaluated for goodness of fit within the model (\S \ref{ssec:irtbackground}). If an item was deemed to fit the ICC poorly or to give rise to local dependence, it was removed for violating model assumptions. Furthermore, if the ICC of an item was too flat, it was removed for low discriminating power between ability levels. The model was then refit with the remaining items. This iterative process continued until no item could be removed (2 to 6 iterations depending on how many items were removed from each set). 

The remaining items make up our final test set ($GS_{IRT}$), which is a calibrated scale of ability to correctly identify the relationship between the two sentence pairs. Parameters of these items were estimated as part of the IRT model and the set of items can be used as an evaluation scale to estimate ability of test-takers or RTE systems. We used the \textit{mirt} R package \cite{chalmers_mirt:_2015} for our analyses.

\section{Results}

\subsection{Response Statistics}

\begin{table}[t]
\small
\centering
\begin{tabular}{|p{3cm}ccc|}
\hline & \bf 4GS & \bf 5GS & \bf Overall \\ \hline
Pairs with majority agreement & 95.6\% & 96.7\% & 96.1\% \\ \hline
Pairs with supermajority agreement & 61.1\% & 82.2\% & 71.7\% \\ \hline
% Individual Label = \newline gold label & 73.2\% & 82.3\% & 77.7\% \\ \hline
% New gold label = \newline original gold label & 81.1\% & 93.3\% & 87.2\% \\ \hline
\end{tabular}
\caption{Summary statistics from the AMT HITs.}\label{tab:amtres}
\end{table}

Table \ref{tab:amtres} lists key statistics from the AMT HITs. Most of the sampled sentence pairs resulted in a gold standard label being identified via a majority vote. Due to the large number of individuals providing labels during the HIT, we also wanted to see if a gold standard label could be determined via a two-thirds supermajority vote. We found that 28.3\% of the sentence pairs did not have a supermajority gold label. This highlights the ambiguity associated with identifying entailment.

We believe that the items selected for analysis are appropriate for our task in that we chose high-quality items, where at least 4 annotators selected the same label, indicating a strong level of agreement (Section \ref{ssec:itemselection}). We argue that our sample is a high-quality portion of the dataset, and further analysis of items where the gold-standard label was only selected by 3 annotators originally would result in lower levels of agreement. 

Table \ref{tab:snlires} shows that the level of agreement as measured by the Fleiss' $\kappa$ score is much lower when the number of annotators is increased, particularly for the 4GS set of sentence pairs, as compared to scores noted in \newcite{bowman_large_2015}. The decrease in agreement is particularly large with regard to contradiction. This could occur for a number of reasons. Recognizing entailment is an inherently difficult task, and classifying a correct label, particularly for contradiction and neutral, can be difficult due to an individual's interpretation of the sentences and assumptions that an individual makes about the key facts of each sentence (e.g. coreference). It may also be the case that the individuals tasked with creating the sentence pairs on AMT created sentences that appeared to contradict a premise text, but can be interpreted differently given a different context.

\begin{table}[t]
\small
\centering
\begin{tabular}{|lccc|}
\hline \bf Fleiss' $\kappa$ & \bf 4GS & \bf 5GS & \bf Bowman et al. 2015  \\ \hline
Contradiction & 0.37 & 0.59 & 0.77  \\	
Entailment & 0.48 & 0.63 & 0.72  \\
Neutral & 0.41 & 0.54 & 0.6  \\
Overall & 0.43 & 0.6 & 0.7  \\
\hline
\end{tabular}
\caption{Comparison of Fleiss' $\kappa$ scores with scores from SNLI quality control sentence pairs.}\label{tab:snlires}
\end{table}

Before fitting the IRT models we performed a visual inspection of the 180 sentence pairs and removed items clearly not suitable for an evaluation scale due to syntactic or semantic discrepancies. For example item 10 in Table \ref{tab:examples} was removed from the 5GS contradiction set for semantic reasons. While many people would agree that the statement is a contradiction due to the difference between football and soccer, individuals from outside the U.S. would possibly consider the two to be synonyms and classify this as entailment. Six such pairs were identified and removed from the set of 180 items, leaving 174 items for IRT model-fitting.

\subsection{IRT Evaluation}
\label{ssec:irtevaluation}

\subsubsection{IRT Models}

We used the methods described in Section \ref{ssec:stats} to build IRT models to scale performance according to the RTE task. For both 4GS and 5GS items three factors were identified, each related to items for the three $GS_{RTE}$ labels (entailment, contradiction, neutral). This suggests that items with the same $GS_{RTE}$ label within each set defines a separate ability. In the subsequent steps, items with different labels were analyzed separately. After analysis, we were left with a subset of the 180 originally selected items. Refer to Table 1 for examples of the retained and removed items based on the IRT analysis. We retained 124 of the 180 items (68.9\%). We were able to retain more items from the 5GS datasets (76 out of 90 - 84\%) than from the 4GS datasets (48 out of 90 - 53.5\%). Items that measure contradiction were retained at the lowest rate for both 4GS and 5GS datasets (66\% in both cases). For the 4GS entailment items, our analysis found that a one-factor model did not fit the data, and a two-factor model failed to yield an interpretable loading pattern after rotation. We were unable to build an IRT model that accurately modeled ability to recognize entailment with the obtained response patterns. As a result, no items from the 4GS entailment set were retained.

\begin{figure}[t]
\centering
\includegraphics[width=0.8\columnwidth]{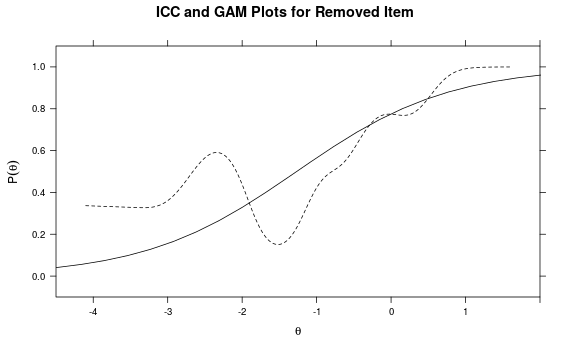}
\caption{Estimated (solid) and actual (dotted) response curves for a removed item.}
\label{fig:iccandgam}
\end{figure}

\begin{figure}[t]
\begin{tabular}{ll}
\includegraphics[width=0.9\columnwidth]{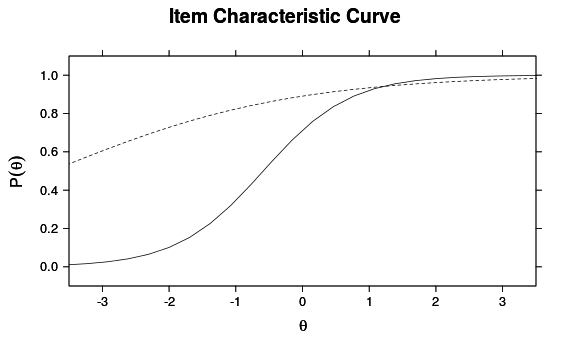}
\end{tabular}
\caption{ICCs for retained (solid) and removed (dotted) items.}
\label{fig:stackediccs}
\end{figure}

Figure \ref{fig:iccandgam} plots the empirical spline-smoothed ICC of one item (Table \ref{tab:examples}, item 9) with its estimated response curve. The ICC is not continuously increasing, and thus a logistic function is not appropriate. This item was spotted for poor item fit and removed. Figure \ref{fig:stackediccs} shows a comparison between the ICC plot of a retained item (Table \ref{tab:examples}, item 4) and the ICC of a removed item (Table \ref{tab:examples}, item 8). Note that the removed item has an ICC that is very flat between -3 and 3. This item cannot discriminate individuals at any common level of ability and thus is not useful. 

The items retained for each factor can be considered as an evaluation scale that measures a single ability of an individual test-taker. As each factor is associated with a separate gold-standard label, each factor ($\theta$) is a person's ability to correctly classify the relationship between the text and hypothesis for one such label (e.g. entailment). 

\subsubsection{Item Parameter Estimation}

Parameter estimates of retained items for each label are summarized in Table \ref{tab:itemparams}, and show that all parameters fall within reasonable ranges. All retained items have 2PL ICCs, suggesting no significant guessing. Difficulty parameters of most items are negative, suggesting that an average AMT user has at least 50\% chance to answer these items correctly. Although some minimum difficulties are quite low for standard ranges for a human population, the low range of item difficulty is appropriate for the evaluation of NLP systems. Items in each scale have a wide range of difficulty and discrimination power.

\begin{table}[t]
\small
\centering
\begin{tabular}{|lp{1cm}p{1cm}p{1cm}p{1cm}|}
\hline 
\bf Item Set & \bf Min. Difficulty & \bf Max. Difficulty & \bf Min. Slope & \bf Max. Slope  \\ \hline
\bf 5GS & & & & \\ \hline
Contradiction & -2.765 & 0.704 & 0.846 & 2.731 \\
Entailment & -3.253 & -1.898 & 0.78 & 2.61 \\ 
Neutral & -2.082 & -0.555 & 1.271 & 3.598 \\ \hline
\bf 4GS & & & & \\ \hline
Contradiction & -1.829 & 1.283 & 0.888 & 2.753 \\ 
Neutral & -2.148 & 0.386 & 1.133 & 3.313 \\ \hline
\end{tabular}
\caption{Parameter estimates of the retained items}\label{tab:itemparams}
\end{table}

\begin{figure}[t]
\centering
\includegraphics[width=0.6\columnwidth]{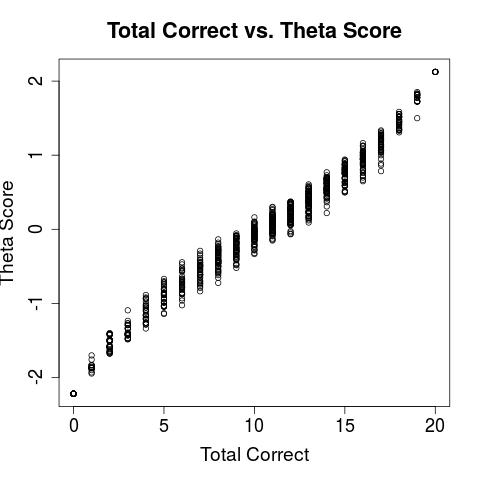}
\caption{Plot of total correct answers vs. IRT scores.}
\label{fig:correctvsIRT}
\end{figure}

With IRT we can use the heterogeneity of items to properly account for such differences in the estimation of a test-taker's ability. Figure \ref{fig:correctvsIRT} plots the estimated ability of each AMT user from IRT against their total number of correct responses to the retained items in the 4GS contradiction item set. The two estimates of ability differ in many aspects. First, test-takers with the same total score may differ in their IRT score because they have different response patterns (i.e. they made mistakes on different items), showing that IRT is able to account for differences among items. Second, despite a rough monotonic trend between the two scores, people with a higher number of correct responses may have a lower ability estimate from IRT. 

We can extend this analysis to the case of RTE systems, and use the newly constructed scales to evaluate RTE systems. A system could be trained on an existing dataset and then evaluated using the retained items from the IRT models to estimate a new ability score. This score would be a measurement of how well the system performed with respect to the human population used to fit the model. With this approach, larger sections of datasets can be devoted to training, with a small portion held out to build an IRT model that can be used for evaluation.

\subsubsection{Application to an RTE System}

As a demonstration, we evaluate the LSTM model presented in \newcite{bowman_large_2015} with the items in our IRT evaluation scales. 
In addition to the theta scores, we calculate accuracy for the binary classification task of identifying the correct label for all items eligible for each subset in Table 5 (e.g. all test items where 5 of 5 annotators labeled the item as \textit{entailment} for 5GS). 
Note that these accuracy metrics are for subsets of the SNLI test set used for binary classifications and therefore do not compare with the standard SNLI test set accuracy measures. 

\begin{table}[t]
\small
\centering
\begin{tabular}{|rcc|p{1.25cm}|}
\hline 
\bf Item Set & \bf Theta Score & \bf Percentile & \bf Test Acc.\\ \hline
\bf 5GS & & & \\ \hline
Entailment & -0.133 & 44.83\% & 96.5\%\\ 
Contradiction & 1.539 & 93.82\% & 87.9\%\\ 
Neutral & 0.423 & 66.28\% & 88\%\\ \hline
\bf 4GS & & & \\ \hline
Contradiction & 1.777 & 96.25\% & 78.9\%\\ 
Neutral & 0.441 & 67\% & 83\%\\ \hline
\end{tabular}
\caption{Theta scores and area under curve percentiles for LSTM trained on SNLI and tested on $GS_{IRT}$. We also report the accuracy for the same LSTM tested on all SNLI quality control items (see Section \ref{ssec:itemselection}). All performance is based on binary classification for each label.}\label{tab:testrun}
\end{table}

The theta scores from IRT in Table \ref{tab:testrun} show that, compared to AMT users, the system performed well above average for contradiction items compared to human performance, and performed around the average for entailment and neutral items. 
For both the neutral and contradiction items, the theta scores are similar across the 4GS and 5GS sets, whereas the accuracy of the more difficult 4GS items is consistently lower. 
This shows the advantage of IRT to account for item characteristics in its ability estimates. 
A similar theta score across sets indicates that we can measure the ``ability level'' regardless of whether the test set is easy or hard. 
Theta score is a consistent measurement, compared to accuracy which varies with the difficulty of the dataset. 

The theta score and accuracy for 5GS entailment show that high accuracy does not necessarily mean that performance is above average when compared to human performance. However, theta score is not meant to contradict accuracy score, but to provide a better idea of system performance compared against a human population. The theta scores are a result of the IRT model fit using human annotator responses and provide more context about the system performance than an accuracy score can alone. If accuracy is high and theta is close to 0 (as is the case with 5GS entailment), we know that the performance of RTE is close to the average level of the AMT user population and that 5GS entailment test set was ``easy'' to both. Theta score and percentile are intrinsically in reference to human performance and independent of item difficulty, while accuracy is intrinsically in reference to a specific set of items.

\section{Discussion and Future Work}

As NLP systems have become more sophisticated, sophisticated methodologies are required to compare their performance. One approach to create an intelligent gold standard is to use IRT to build models to scale performance on a small section of items with respect to the tested population. IRT models can identify dataset items with different difficulty levels and discrimination powers based on human responses, and identify items that are not appropriate as scale items for evaluation. The resulting small set of items can be used as a scale to score an individual or NLP system. This leaves a higher percentage of a dataset to be used in the training of the system, while still having a valuable metric for testing.

IRT is not without its challenges. A large population is required to provide the initial responses in order to have enough data to fit the models; however, crowdsourcing allows for the inexpensive collection of large amounts of data. An alternative methodology is Classical Test Theory, which has its own limitations, in particular that it is test-centric, and cannot provide information for individual items. 

We have introduced Item Response Theory from psychometrics as an alternative method for generating gold-standard evaluation datasets. Fitting IRT models allows us to identify a set of items that when taken together as a test set, can provide a meaningful evaluation of NLP systems with the different difficulty and discriminating characteristics of the items taken into account. We demonstrate the usefulness of the IRT-generated test set by showing that high accuracy does not necessarily indicate high performance when compared to a population of humans. 

Future work can adapt this analysis to create evaluation mechanisms for other NLP tasks. The expectation is that systems that perform well using a standard accuracy measure can be stratified based on which types of items they perform well on. High quailty systems should also perform well when the models are used together as an overall test of ability. This new evaluation for NLP systems can lead to new and innovative methods that can be tested against a novel benchmark for performance, instead of gradually incrementing on a classification accuracy metric.

% uncomment for final submission
\section*{Acknowledgments}
We thank the AMT Turkers who completed our annotation task.
We would like to also thank the anonymous reviewers for their insightful comments.

This work was supported in part by the HSR\&D award IIR 1I01HX001457 from the United States Department of Veterans Affairs (VA). We also acknowledge the support of HL125089 from the National Institutes of Health.  
%grant HL125089 from the National Institutes of Health (NIH). 
%We also acknowledge the support from the United States Department of Veterans Affairs (VA) through Award 1I01HX001457.
This work was also supported in part by the Center for Intelligent Information Retrieval. The contents of this paper do not represent the views of CIIR, NIH, VA, or the United States Government

\bibliography{jlalor}
\bibliographystyle{emnlp2016}

\end{document}